\documentclass[runningheads]{llncs}
\usepackage[T1]{fontenc}
\usepackage{graphicx,verbatim}
\usepackage{tabularx}
\usepackage{booktabs}
\usepackage{amsmath}
\usepackage{float}
\usepackage{amsfonts,amssymb}
\usepackage{array}
\usepackage{siunitx}
\usepackage{hyperref}
\begin{document}
\title{Hi-GaTA: Hierarchical Gated Temporal Aggregation Adapter for Surgical Video Report Generation}
%

\author{
    Kedi Sun\inst{1} \and 
    Chaohui Dang\inst{1} \and 
    Yue Feng\inst{2} \and 
    James Glasbey\inst{3} \and 
    Theodoros N. Arvanitis\inst{1} \and 
    Le Zhang\inst{1}
}

\authorrunning{K. Sun et al.} 

\institute{
    School of Engineering, College of Engineering and Physical Sciences, University of Birmingham, Birmingham, UK \\
    \and
    School of Computer Science, University of Birmingham, Birmingham, UK \\
    \and
    Department of Applied Health Sciences, University of Birmingham, Birmingham, UK\\
    \email{kxs1163@student.bham.ac.uk; l.zhang.16@bham.ac.uk}
}
  
\maketitle              
\begin{abstract}
Automated, clinician-grade assessment reports for surgical procedures could reduce documentation burden and provide objective feedback, yet remain challenging 
due to the difficulty of aligning dense spatio-temporal video representations with language-based reasoning and the scarcity of high-quality, privacy-preserving datasets. 
To address this gap, we establish a benchmark comprising 214 high-quality simulated surgical videos\footnote{Operations on physical phantoms by human surgeons} paired with surgeon-authored evaluation reports. Building on this resource, we propose a Perception–Alignment–Reasoning framework for surgical video report generation, featuring \textit{Hi-GaTA}, a novel lightweight temporal adapter that efficiently compresses long video sequences into compact, LLM-compatible visual prefix tokens through short-to-long-range temporal aggregation.
For robust visual perception, we pretrain \textit{Sur40k}, a surgical-specific ViViT-style video encoder on 40,000 minutes of public surgical videos to capture fine-grained spatio-temporal procedural priors. 
\textit{Hi-GaTA} employs a temporal pyramid with text-conditioned dual cross-attention, and improves multi-scale consistency through cross-level gated fusion and an increasing-depth strategy.
Finally, we fine-tune the LLM backbone using LoRA to enable coherent and stylistically consistent surgical report generation under limited supervision. 
Experiments show our approach achieves the best overall performance, with consistent gains over strong Multimodal Marge Language Model (MLLM) baselines. 
Ablation studies further validate the effectiveness of each proposed component. \textit{Code will be available after review}.

\end{abstract}

\section{Introduction}
Recent advances in computer assisted-intervention (CAI) have seen profound success in surgical phase recognition~\cite{yuan2024hecvl} and atomic action detection~\cite{nwoye2022rendezvous}. However, these perception-level tasks do not meet a pressing clinical need: automated, interpretable assessment of procedural quality. 
Currently, conventional surgical skills assessment and reporting rely primarily on manual evaluation by senior experts, which is inherently labor-intensive, plagued by high inter-expert linguistic variability, and lacking in objective standardization\cite{maier2017surgical}. Moving from action recognition to automated, clinician-grade report generation and skill assessment would significantly reduce the documentation burden and provide fine-grained, objective feedback for surgical training and clinical quality assurance.

Despite the rapid evolution of Multimodal Large Language Models (MLLMs) in general medical visual understanding~\cite{li2023llava}~\cite{li2025mu}, their application to surgical report generation remains highly challenging. First, surgical procedures are inherently long-range temporal processes characterized by complex dependencies between global workflow phases and fine-grained local actions~\cite{twinanda2016endonet}. Existing MLLM frameworks struggle to capture these multi-scale temporal dynamics~\cite{li2024mvbench}~\cite{li2024vit3d}, limiting their ability to capture the relationship between subtle movements and overall surgical progression. Second, directly injecting dense video features into Large Language Models (LLM) incurs significant computational redundancy~\cite{song2024moviechat}, simultaneously overwhelming the LLMs with uninformative noise~\cite{zhao2025video}, which can impair and degrade its reasoning and comprehension capabilities. Finally, progress in this area is constrained by the scarcity of high-quality annotated datasets~\cite{wang2025endochat}, which are critical for learning the complexity of surgical procedures~\cite{de2025scaling,yuan2024procedure} and handling the high linguistic variability of surgical reports encountered in real clinical practice.

In this work, we present a holistic framework for surgical video report generation, developed through a close collaboration between university researchers and clinical experts. \textbf{Our contributions:}
\textbf{(1)} We introduce the novel task of surgical video report generation, tackled by a tailored Perception--Alignment--Reasoning MLLM framework. As its foundational perception module, we propose \textit{Sur40k}, a self-supervised video encoder specifically designed to capture fine-grained spatio-temporal patterns and procedural priors. 
\textbf{(2)} As a foundation for this emerging task, we introduce a benchmark dataset comprising 214 high-quality simulation surgical videos, each paired with detailed evaluation reports authored by professional surgeons.
\textbf{(3)} To bridge the semantic gap between dense video representations and linguistic reasoning, we further introduce \textit{Hi-GaTA} (Hierarchical Gated Temporal Aggregation), a lightweight and plug-and-play temporal adapter. \textit{Hi-GaTA} employs a query-based temporal pyramid that iteratively refines learnable queries via dual cross-attention and cross-level gating to distill multi-scale procedural dynamics into compact visual prefix tokens. Finally, to enable clinically coherent and stylistically consistent report generation, we adapt the LLM backbone using Low-Rank Adaptation (LoRA)~\cite{hu2022lora}, allowing the model to reason over complex surgical videos without full-parameter fine-tuning.

\section{Method}

\subsection{Task Formulation}
We investigate surgical video report generation as a translational advancement of surgical perception tailored for clinical application. Given a surgical video \textbf{V} depicting a complete simulated surgical procedure and the prompt text $\mathbf{X}_{\text{inst}}$, the goal is to generate an interpretable surgical report \textbf{Y} that summarizes overall assessment of the surgery and skill-relevant events guided by prompt $\mathbf{X}_{\text{inst}}$.

\textbf{Video feature extraction via \textit{Sur40k} Encoder:}
We segment the input video $\mathbf{V}$ into non-overlapping fixed-length clips. Each clip is encoded by \textit{Sur40k} (based on factorized ViViT architecture~\cite{arnab2021vivit}) to process 3D tubelets via cascaded spatial-temporal transformers. For each clip, we extract the final temporal class token $[\text{CLS}]_i$ as its compact feature representation $h_i \in \mathbb{R}^D$, where $D$ denotes the encoder dimension. Consequently, concatenating these representations along the temporal axis encodes the entire video as a temporal sequence of window-level embeddings $H = [h_1, \dots, h_N]$. This sequence serves as the input for the subsequent downstream temporal adapter and LLM-based surgical report generation.

\textbf{From video alignment to report generation:}
Let $H$ be the sequence of window embeddings. The \textit{Hi-GaTA} module hierarchically aggregates temporal features to transform long-range video representations into $N_p$ visual prefix tokens, yielding a compact representation $\mathbf{P} \in \mathbb{R}^{N_p \times D_h}$ in the LLM's embedding space. Subsequently, the model employs standard causal decoding~\cite{vaswani2017attention} to auto-regressively generate the report \textbf{Y}.
During training, we adopt a two-stage strategy: we first keep the LLM backbone frozen and train only the \textit{Hi-GaTA} modules, and then fine-tune the LLM with LoRA while keeping \textit{Hi-GaTA} frozen to enable coherent report generation under limited surgical-report supervision.

An overview of the proposed Perception–Alignment–Reasoning pipeline is provided in Fig.~1. Specifically, we first encode the long-horizon video into a sequence of window embeddings using \textit{Sur40k}, then convert them into a compact set of LLM-compatible visual prefix tokens via \textit{Hi-GaTA}, and finally concatenate the visual prefix tokens with the prompt tokens for surgical report generation.

\begin{figure}[t]
    \centering
    \includegraphics[width=\linewidth]{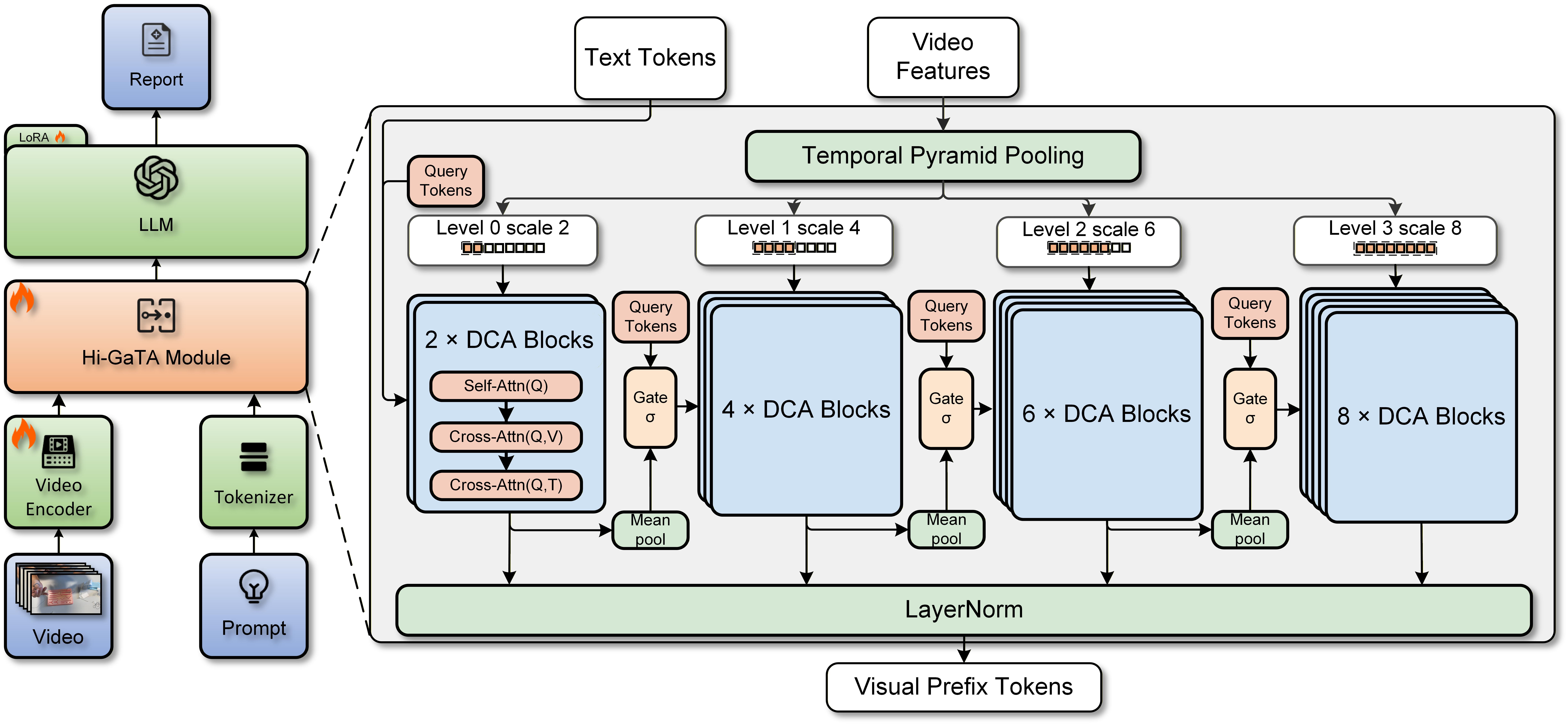}
    \caption{Overview of our proposed method. Left: The Perception--Alignment--Reasoning pipeline for surgical video report generation. Right: Detailed architecture of the \textit{Hi-GaTA} module.}
    \label{fig:higata_framework}
\end{figure}

\subsection{Surgical-Specific Video Encoder (\textit{Sur40k})}
To establish a robust visual foundation for report generation, we employ a self-supervised learning framework to pre-train \textit{Sur40k}. The core principle is to maximize the mutual information between different augmented views of the same video clip while minimizing it for distinct clips. Specifically, given a raw video, we randomly decode a temporal window of $W$ seconds and uniformly sample $F$ frames. We then apply a stochastic augmentation pipeline twice independently to generate two correlated views, $\mathbf{v}^{(1)}$ and $\mathbf{v}^{(2)}$. 
This pipeline applies random resized cropping, color jittering, and grayscale to learn representation invariance against low-level visual perturbations. These views are processed by the ViViT backbone to extract high-dimensional representations, which are then mapped to a lower-dimensional latent space via a non-linear projection head (comprising linear layers with GELU activation and LayerNorm). The resulting projections, $\mathbf{z}^{(1)}, \mathbf{z}^{(2)} \in \mathbb{R}^{D}$ are $\ell_2$-normalized. We optimize the encoder using a symmetric InfoNCE objective~\cite{oord2018representation}, which treats $(\mathbf{z}^{(1)}_i, \mathbf{z}^{(2)}_i)$ as a positive pair and all other augmented views in the minibatch as negatives. The loss function is defined as:
\begin{equation}
\mathcal{L}_{\mathrm{NCE}} = \frac{1}{2} \left[ \mathrm{CE}\!\left( \frac{\mathbf{Z}^{(1)} {\mathbf{Z}^{(2)}}^\top}{\tau}, \, \mathbf{I} \right) + \mathrm{CE}\!\left( \frac{\mathbf{Z}^{(2)} {\mathbf{Z}^{(1)}}^\top}{\tau}, \, \mathbf{I} \right) \right]
\end{equation}
where $\mathbf{Z}^{(1)}, \mathbf{Z}^{(2)} \in \mathbb{R}^{B \times D}$ denote the batches of embeddings, $\tau$ is a temperature hyperparameter and $\mathrm{CE}(\cdot, \mathbf{I})$ represents the cross-entropy loss with the identity matrix $\mathbf{I}$ as ground-truth labels. This pretraining strategy equips \textit{Sur40k} with robust medical spatio-temporal priors and the capability to precisely capture fine-grained surgical details for the downstream report generation task.

\subsection{\textit{Hi-GaTA}: Hierarchical Gated Temporal Aggregation}
Directly injecting dense long-range video tokens into LLMs is computationally inefficient and prone to information redundancy~\cite{song2024moviechat}. To address this, we propose \textit{Hi-GaTA} (Hierarchical Gated Temporal Aggregation), a lightweight, plug-and-play temporal adapter  to bridge the gap between long video sequences and LLMs. \textit{Hi-GaTA} efficiently compresses surgical video feature sequences into a compact set of LLM-compatible visual prefix tokens while explicitly preserving multi-scale temporal structures. As illustrated in Fig.~1, our module extracts multi-scale video features via a short-to-long-range temporal pyramid and hierarchically aggregates them into text-conditioned query tokens~\cite{li2023blip} through cross-attention.

\textbf{Temporal Pyramid Pooling.}
We first build multi-scale temporal representations using Temporal Pyramid Pooling (TPP)~\cite{wang2016tpp}. For a set of window sizes ${\{w_\ell\}}_{\ell=1}^L$, TPP slides a mean-pooling kernel of width $w_\ell$ over the feature sequence $H$ with stride $\gamma w_\ell$ (stride factor $\gamma \in (0,1]$), producing pooled sequences:
\begin{equation}
    H_\ell = TPP_\ell(H)\in\mathbb{R}^{S_\ell\times D}, \quad \ell=1,\dots,L
\end{equation}
$S_\ell$ is the pooled sequence length at scale $\ell$. In our main setting, we use $L=4$ scales with window sizes $(2,4,6,8)$ and $\gamma=0.5$.

\textbf{Query-Based Hierarchical Aggregation with Text Conditioning.}
For each scale $\ell$, \textit{Hi-GaTA} initializes a set of learnable query tokens $Q_\ell \in \mathbb{R}^{N_q \times D_h}$. To enable cross-modal interaction, we project the pooled visual features $H_\ell$ into the LLM's hidden dimension $D_h$, yielding the visual context $\tilde{H}_\ell \in \mathbb{R}^{S_\ell \times D_h}$. These queries are then iteratively updated by a stack of Dual Cross-Attention (DCA) blocks via: (i) self-attention~\cite{vaswani2017attention} across $Q_\ell$, (ii) cross-attention to the visual context $\tilde{H}_\ell$, and (iii) cross-attention to the prompt text tokens $\mathbf{E}_{\text{inst}}\in\mathbb{R}^{L_p\times D_h}$, where $L_p$ is the sequence length of the prompt. This design explicitly aligns the hierarchical temporal aggregation with the textual prompt.

\textbf{Cross-Level Gating and Increasing-Depth Strategy.}
\textit{Hi-GaTA} introduces two key mechanisms to better align video representations with the LLM's semantic space, enabling the model to capture clinically relevant temporal semantics and produce more faithful, coherent reports.

\textit{Cross-level gated fusion:} To propagate short-range context to long-range, we summarize the previous-level queries and inject them into the current level via a sigmoid gate. 
Given the updated queries from level $\ell\!-\!1$, $Q_{\ell-1}\in\mathbb{R}^{N_q\times D_h}$, we compute a short-range context by mean pooling over the query dimension:
\begin{equation}
c_{\ell-1}=\frac{1}{N_q}\sum_{q=1}^{N_q} Q_{\ell-1}^{(q)} \in \mathbb{R}^{1\times D_h}
\end{equation}
A channel-wise gate is then produced by a lightweight linear projection followed by a sigmoid, where $W_g$ is the learnable weight controlling channel activation:
\begin{equation}
g_\ell=\sigma(W_g c_{\ell-1}) \in \mathbb{R}^{1\times D_h}
\end{equation}
Finally, we inject the gated context into all query tokens at level $\ell$ through a residual update (broadcast along the query dimension):
\begin{equation}
Q_\ell \leftarrow Q_\ell + g_\ell \odot c_{\ell-1}, \qquad Q_\ell\in\mathbb{R}^{N_q\times D_h}
\end{equation}
This gated residual injection transfers short-range fine-grained contexts to support long-range temporal summarization and encourages cross-scale semantic consistency throughout the report generation process.

\textit{Increasing depth with block sharing:} We implement a shared global block pool where the number of executed blocks progressively increases along the hierarchy. This creates a non-uniform compute budget that prioritizes long-range temporal abstraction while maintaining parameter efficiency via weight sharing. Finally, we concatenate and normalize queries from all scales to form visual prefix tokens:
\begin{equation}
    \mathbf{P}=\mathrm{LN}([Q_1;\dots;Q_L])\in\mathbb{R}^{N_p\times D_h}
\end{equation}
In our work, we set $N_q=4$ and $L=4$, resulting in $N_p=16$ visual prefix tokens.

\subsection{Optimization}
We optimize the model across both training stages using a composite objective that combines the auto-regressive generation loss with a regularization term to prevent overfitting and stabilize training. The total loss function is defined as:
\begin{equation}
\mathcal{L} = - \sum_{t=1}^{N_y} \log p(y_t \mid \mathbf{P}, \mathbf{E_{\text{inst}}}, y_{<t}) + \frac{\lambda}{N} \|P\|_F^2
\end{equation}
where $N_{y}$ is the token length of the ground-truth report $Y$. The first term is the Negative Log-Likelihood (NLL), and the second term applies mean-squared regularization on the visual prefix tokens $\mathbf{P}$ with $N$ total elements and $\lambda=0.02$.

\section{Experiments}

\textbf{Datasets and Preprocessing.} 
We introduce a benchmark dataset under a university--hospital collaboration protocol, comprising 214 high-quality surgical procedure videos ranging from 1 to 9 minutes in duration, paired with expert surgeon evaluations. In this benchmark, each sample contains: (i) seven Objective Structured Assessment of Technical Skills (OSATS)~\cite{niitsu2013using} style global rating dimensions and an overall rating scored by the surgeon on a 1--5 Likert scale. (ii) corresponding free-text feedback written as a narrative assessment by the surgeon. To construct robust training targets for the \textit{Hi-GaTA} module, we synthesize a unified reference report for each video by integrating the structured scores with the surgeon's narrative assessments using \textit{GPT-5}, strictly preserving the original linguistic style and clinical nuances of the surgeons.


For the pretraining of \textit{Sur40k}, we curated a dataset comprising 40,000 minutes of authentic surgical video collected from publicly available sources. This enables \textit{Sur40k} to establish robust surgical visual representations via self-supervised learning, providing a solid foundation for the downstream report generation task.

\textbf{Implementation Details.}
All experiments are conducted on a single 40G NVIDIA A100 GPU. For \textit{Sur40k} pretraining, we segment the dataset into 1-minute videos, sampling 16-frame clips from a random 4.0-second window. We optimize the model using AdamW with a batch size of 16 and weight decay of 0.05. The learning rate follows a cosine annealing schedule, decaying from $3 \times 10^{-4}$ to $1 \times 10^{-6}$, while gradient clipping is applied for training stability.

For the downstream report generation task, we train our framework on our established benchmark dataset, we reserve 20 videos for independent testing and partition the remaining 194 into training and validation sets using an 80/20 split. We adopt a two-stage optimization strategy utilizing the AdamW optimizer. In the first stage, we freeze the LLM and train the \textit{Hi-GaTA} module for 30 epochs with a batch size of 8, applying label smoothing of 0.05 and mean-squared prefix regularization. The learning rate follows a cosine schedule with 100 warm-up steps, decays from $1 \times 10^{-5}$. In the second stage, we freeze \textit{Hi-GaTA} and fine-tune the LLM via LoRA, configured with a rank of 8, an alpha scaling factor of 16 and a dropout rate of 0.2. This phase proceeds for 30 epochs with a reduced batch size of 4 to accommodate gradient memory, maintaining the identical cosine schedule with 100 warm-up steps and a peak learning rate of $1 \times 10^{-5}$.

\textbf{Comparison and Evaluation Metrics.}
We conduct controlled comparisons to disentangle the impact of the video encoder and the LLM backbone. Specifically, Table~1 reports two complementary settings: (i) encoder comparison, where we swap the video encoder (Dinov3~\cite{simeoni2025dinov3}, ViViT~\cite{arnab2021vivit}, XCLIP~\cite{ni2022expanding}, and our \textit{Sur40k}) while keeping the LLM(Qwen2.5-1.5B) fixed. (ii) LLM comparison, where we vary the LLM backbone (Gemma2-2B~\cite{team2024gemma}, Llama3.2-1B/3B~\cite{dubey2024llama}, and Qwen2.5-1.5B/3B~\cite{qwen2025qwen25technicalreport}) while keeping the video encoder(\textit{Sur40k}) fixed. In addition, we compare against two strong MLLM baselines, LLaVA-Med-v1.5~\cite{li2023llava} and Qwen2.5-VL-7B~\cite{bai2025qwen2} specifically prompted to follow the target reporting format.

Following standard practice for report generation, we evaluate with a five-dimensional suite capturing both lexical overlap and semantic similarity: BLEU~\cite{papineni2002bleu}, ROUGE-L~\cite{lin2004rouge}, METEOR~\cite{banerjee2005meteor}, MedBERTScore~\cite{gu2021domain}, and CIDEr~\cite{vedantam2015cider}. All metrics are computed between generated reports and  reference reports.

\newcommand{\mstd}[2]{#1$\pm$#2}

\begin{table}[t]
\centering
\caption{Joint comparison of video encoders and LLM backbones (mean $\pm$ std).}
\label{tab:enc_llm}

\begingroup
\fontsize{8}{9}\selectfont
\setlength{\tabcolsep}{2.2pt}
\renewcommand{\arraystretch}{1}

\begin{tabularx}{\linewidth}{@{}l*{5}{>{\centering\arraybackslash}X}@{}}
\toprule
Encoder & BLEU & ROUGE-L & METEOR & MedBERTScore & CIDEr \\
\midrule
Dinov3~\cite{simeoni2025dinov3}                 & \mstd{0.321}{0.106} & \mstd{0.239}{0.108} & \mstd{0.309}{0.102} & \mstd{0.893}{0.013} & \mstd{0.438}{0.258} \\
ViViT~\cite{arnab2021vivit}                  & \mstd{0.343}{0.111} & \mstd{0.266}{0.089} & \mstd{0.336}{0.117} & \mstd{0.895}{0.013} & \mstd{0.264}{0.330} \\
XCLIP~\cite{ni2022expanding}                  & \mstd{0.332}{0.086} & \mstd{0.246}{0.074} & \mstd{0.310}{0.090} & \mstd{0.893}{0.010} & \mstd{0.212}{0.319} \\
\textit{Sur40k} (ours) & \textbf{\mstd{0.414}}{0.167} & \textbf{\mstd{0.365}}{0.106} & \textbf{\mstd{0.416}}{0.134} & \textbf{\mstd{0.906}}{0.017} & \textbf{\mstd{0.675}} {0.951}\\
\bottomrule
\end{tabularx}

\begin{tabularx}{\linewidth}{@{}l*{5}{>{\centering\arraybackslash}X}@{}}
\toprule
LLM Backbone & BLEU & ROUGE-L & METEOR & MedBERTScore & CIDEr \\
\midrule
Gemma2-2B~\cite{team2024gemma}            & \mstd{0.367}{0.085} & \mstd{0.289}{0.088} & \mstd{0.355}{0.091} & \mstd{0.891}{0.011} & \mstd{0.259}{0.257} \\
Llama3.2-3B~\cite{dubey2024llama}          & \mstd{0.374}{0.080} & \mstd{0.301}{0.136} & \mstd{0.369}{0.110} & \mstd{0.893}{0.019} & \mstd{0.214}{0.230} \\
Llama3.2-1B~\cite{dubey2024llama}          & \mstd{0.364}{0.120} & \mstd{0.310}{0.151} & \mstd{0.374}{0.153} & \mstd{0.896}{0.017} & \mstd{0.303}{0.317} \\
Qwen2.5-3B~\cite{qwen2025qwen25technicalreport}           & \mstd{0.346}{0.132} & \mstd{0.284}{0.107} & \mstd{0.338}{0.132} & \mstd{0.896}{0.014} & \mstd{0.331}{0.448} \\
Qwen2.5-1.5B~\cite{qwen2025qwen25technicalreport}        & \textbf{\mstd{0.414}}{0.167} & \textbf{\mstd{0.365}}{0.106} & \textbf{\mstd{0.416}}{0.134} & \textbf{\mstd{0.906}}{0.017} & \textbf{\mstd{0.675}}{0.951} \\
\bottomrule
\end{tabularx}

\par\endgroup
\end{table}

\begin{table}[t]
\centering
\caption{Ablation of \textit{Hi-GaTA} and comparison with MLLM baselines (mean $\pm$ std).}
\label{tab:ablation_MLLM}

\begingroup
\fontsize{8}{9}\selectfont
\setlength{\tabcolsep}{2.2pt}
\renewcommand{\arraystretch}{1}
\begin{tabularx}{\linewidth}{@{}l*{5}{>{\centering\arraybackslash}X}@{}}
\toprule
Model & BLEU & ROUGE-L & METEOR & MedBERTScore & CIDEr \\
\midrule
LLaVA-Med-v1.5-7B~\cite{li2023llava}   & \mstd{0.127}{0.082} & \mstd{0.141}{0.041} & \mstd{0.142}{0.060} & \mstd{0.802}{0.010} & \mstd{0.009}{0.031} \\
Qwen2.5-VL-7B~\cite{bai2025qwen2}      & \mstd{0.184}{0.047} & \mstd{0.132}{0.033} & \mstd{0.156}{0.031} & \mstd{0.803}{0.011} & \mstd{0.057}{0.067} \\
w/o \textit{Hi-GaTA}   & \mstd{0.184}{0.057} & \mstd{0.085}{0.034} & \mstd{0.165}{0.055} & \mstd{0.840}{0.013} & \mstd{0.033}{0.073} \\
Only Depth-Increasing & \mstd{0.345}{0.013} & \mstd{0.272}{0.012} & \mstd{0.339}{0.124} & \mstd{0.896}{0.016} & \mstd{0.305}{0.394} \\
Only Gated fusion         & \mstd{0.364}{0.107} & \mstd{0.291}{0.094} & \mstd{0.362}{0.113} & \mstd{0.898}{0.013} & \mstd{0.338}{0.437} \\
\textit{Hi-GaTA} full       &  \textbf{\mstd{0.414}}{0.167} & \textbf{\mstd{0.365}}{0.106} & \textbf{\mstd{0.416}}{0.134} & \textbf{\mstd{0.906}}{0.017} & \textbf{\mstd{0.675}}{0.951} \\
\bottomrule
\end{tabularx}

\par\endgroup
\end{table}

\textbf{Results Analysis.}
Our quantitative results are summarized in Table~1 and Table~2. Table~1 demonstrates the critical role of surgical-domain perception in processing long-horizon videos. Specifically, replacing \textit{Sur40k} with general-purpose encoders consistently degrades performance, suggesting that procedural priors from large-scale pretraining provide essential visual grounding for surgical report generation. Furthermore, performance is not primarily driven by LLM scaling: a compact backbone (Qwen2.5-1.5B) achieves the strongest overall scores when paired with \textit{Sur40k}. The large CIDEr variance stems from the high linguistic variability of expert narratives and strict n-gram matching, which heavily penalizes clinically valid synonymous descriptions. Nevertheless, our consistently high and stable MedBERTScore confirms that the model robustly captures true clinical semantics despite these lexical variations.

Table~2 further shows that our approach consistently outperforms strong MLLM baselines (Qwen2.5-VL-7B and LLaVA-Med-v1.5-7B), demonstrating the superiority of explicit temporal aggregation for long-horizon surgical videos. While standard MLLMs struggle to retain long-term dependencies, our method effectively mitigates such information decay. Ablation results validate our core designs: removing either the cross-level gated fusion or the increasing-depth strategy causes significant performance drops. Specifically, gated fusion enhances cross-scale semantic consistency, whereas increasing depth facilitates holistic procedural understanding. Finally, simultaneous improvements in semantic and lexical metrics verify that \textit{Hi-GaTA} grounds narratives in specific procedural details while strictly adhering to professional clinical reporting standards.

Qualitative examples are shown in Fig.~2. Compared with Qwen2.5-VL-7B and LLaVA-Med-v1.5-7B, our \textit{Hi-GaTA} approach produces more clinically accurate and comprehensive surgical assessment report, better matching the ground-truth reference, consistent with the semantic improvements in Table~2.

\begin{figure}[t]
    \centering
    \includegraphics[width=\linewidth]{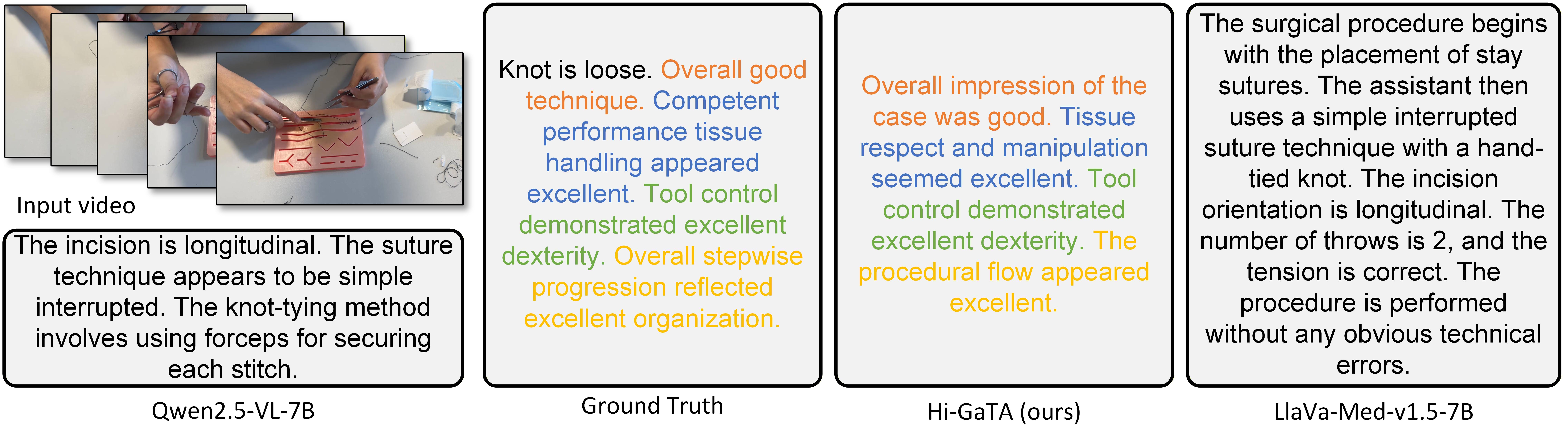}
    \caption{Qualitative comparison of generated reports. Our \textit{Hi-GaTA} approach produces more clinically accurate and comprehensive descriptions than LLaVA-Med-v1.5-7B~\cite{li2023llava} and Qwen2.5-VL-7B~\cite{bai2025qwen2}, closer to the ground truth.}
    \label{fig:qualitative}
\end{figure}

\section{Conclusion}
In this work, we presented \textit{Hi-GaTA}, a novel hierarchical gated temporal adapter tailored for long-horizon surgical video report generation. Unlike standard MLLM paradigms that struggle with temporal complexity, \textit{Hi-GaTA} explicitly models procedural dynamics through short-to-long-range aggregation, effectively compressing dense video features into compact, LLM-compatible visual prefix tokens. 
We validated the efficacy of our framework on our established benchmark dataset comprising 214 expert-annotated surgical videos, which addresses the critical scarcity of high-quality supervision in this field. Furthermore, our results highlight the critical role of domain-specific perception. We demonstrate that \textit{Sur40k}, pretrained on 40,000 minutes of surgical video, establishes a robust visual foundation that significantly boosts the performance of the \textit{Hi-GaTA} approach. In future work, we aim to validate our framework across multi-center datasets covering diverse procedures and investigate expert-in-the-loop optimization to further enhance the clinical utility and safety of automated surgical reporting.

    

%
%
%
%
\bibliographystyle{splncs04} 
\bibliography{ref}

@article{li2024vit3d,
  title={Vit3d alignment of llama3: 3d medical image report generation},
  author={Li, Siyou and Xu, Beining and Luo, Yihao and Nie, Dong and Zhang, Le},
  journal={arXiv preprint arXiv:2410.08588},
  year={2024}
}

@inproceedings{li2025mu,
  title={$\mu$ 2 Tokenizer: Differentiable Multi-Scale Multi-Modal Tokenizer for Radiology Report Generation},
  author={Li, Siyou and Qin, Pengyao and Wu, Huanan and Nie, Dong and Thirunavukarasu, Arun J and Yu, Juntao and Zhang, Le},
  booktitle={International Conference on Medical Image Computing and Computer-Assisted Intervention},
  pages={3--12},
  year={2025},
  organization={Springer}
}

@article{yuan2024procedure,
  title={Procedure-aware surgical video-language pretraining with hierarchical knowledge augmentation},
  author={Yuan, Kun and Navab, Nassir and Padoy, Nicolas and others},
  journal={Advances in Neural Information Processing Systems},
  volume={37},
  pages={122952--122983},
  year={2024}
}

@inproceedings{yuan2024hecvl,
  title={Hecvl: Hierarchical video-language pretraining for zero-shot surgical phase recognition},
  author={Yuan, Kun and Srivastav, Vinkle and Navab, Nassir and Padoy, Nicolas},
  booktitle={International Conference on Medical Image Computing and Computer-Assisted Intervention},
  pages={306--316},
  year={2024},
  organization={Springer}
}

@article{de2025scaling,
  title={Scaling up self-supervised learning for improved surgical foundation models},
  author={de Jong, Ronald and Carolus, HJ and Franciscus, HA and van Jaarsveld, Romy C and van Hillegersberg, Richard and Josien, PW and de With, Peter HN and al Khalil, Yasmina and van Der Sommen, Fons and others},
  journal={Medical Image Analysis},
  pages={103873},
  year={2025},
  publisher={Elsevier}
}

@article{nwoye2022rendezvous,
  title={Rendezvous: Attention mechanisms for the recognition of surgical action triplets in endoscopic videos},
  author={Nwoye, Chinedu Innocent and Yu, Tong and Gonzalez, Cristians and Seeliger, Barbara and Mascagni, Pietro and Mutter, Didier and Marescaux, Jacques and Padoy, Nicolas},
  journal={Medical Image Analysis},
  volume={78},
  pages={102433},
  year={2022},
  publisher={Elsevier}
}

@article{bai2025qwen2,
  title={Qwen2. 5-vl technical report},
  author={Bai, Shuai and Chen, Keqin and Liu, Xuejing and Wang, Jialin and Ge, Wenbin and Song, Sibo and Dang, Kai and Wang, Peng and Wang, Shijie and Tang, Jun and others},
  journal={arXiv preprint arXiv:2502.13923},
  year={2025}
}

@inproceedings{li2023blip,
  title={Blip-2: Bootstrapping language-image pre-training with frozen image encoders and large language models},
  author={Li, Junnan and Li, Dongxu and Savarese, Silvio and Hoi, Steven},
  booktitle={International conference on machine learning},
  pages={19730--19742},
  year={2023},
  organization={PMLR}
}

@article{li2023llava,
  title={Llava-med: Training a large language-and-vision assistant for biomedicine in one day},
  author={Li, Chunyuan and Wong, Cliff and Zhang, Sheng and Usuyama, Naoto and Liu, Haotian and Yang, Jianwei and Naumann, Tristan and Poon, Hoifung and Gao, Jianfeng},
  journal={Advances in Neural Information Processing Systems},
  volume={36},
  pages={28541--28564},
  year={2023}
}

@inproceedings{arnab2021vivit,
  title={Vivit: A video vision transformer},
  author={Arnab, Anurag and Dehghani, Mostafa and Heigold, Georg and Sun, Chen and Lu{\v{c}}i{\'c}, Mario and Schmid, Cordelia},
  booktitle={Proceedings of the IEEE/CVF international conference on computer vision},
  pages={6836--6846},
  year={2021}
}

@article{wang2016tpp,
  title={Temporal pyramid pooling-based convolutional neural network for action recognition},
  author={Wang, Peng and Cao, Yuanzhouhan and Shen, Chunhua and Liu, Lingqiao and Shen, Heng Tao},
  journal={IEEE Transactions on Circuits and Systems for Video Technology},
  volume={27},
  number={12},
  pages={2613--2622},
  year={2016},
  publisher={IEEE}
}

@article{hu2022lora,
  title={Lora: Low-rank adaptation of large language models.},
  author={Hu, Edward J and Shen, Yelong and Wallis, Phillip and Allen-Zhu, Zeyuan and Li, Yuanzhi and Wang, Shean and Wang, Lu and Chen, Weizhu and others},
  journal={ICLR},
  volume={1},
  number={2},
  pages={3},
  year={2022}
}

@inproceedings{papineni2002bleu,
  title={Bleu: a method for automatic evaluation of machine translation},
  author={Papineni, Kishore and Roukos, Salim and Ward, Todd and Zhu, Wei-Jing},
  booktitle={Proceedings of the 40th annual meeting of the Association for Computational Linguistics},
  pages={311--318},
  year={2002}
}

@inproceedings{lin2004rouge,
  title={Rouge: A package for automatic evaluation of summaries},
  author={Lin, Chin-Yew},
  booktitle={Text summarization branches out},
  pages={74--81},
  year={2004}
}

@inproceedings{banerjee2005meteor,
  title={METEOR: An automatic metric for MT evaluation with improved correlation with human judgments},
  author={Banerjee, Satanjeev and Lavie, Alon},
  booktitle={Proceedings of the acl workshop on intrinsic and extrinsic evaluation measures for machine translation and/or summarization},
  pages={65--72},
  year={2005}
}

@inproceedings{vedantam2015cider,
  title={Cider: Consensus-based image description evaluation},
  author={Vedantam, Ramakrishna and Lawrence Zitnick, C and Parikh, Devi},
  booktitle={Proceedings of the IEEE conference on computer vision and pattern recognition},
  pages={4566--4575},
  year={2015}
}

@article{vaswani2017attention,
  title={Attention is all you need},
  author={Vaswani, Ashish and Shazeer, Noam and Parmar, Niki and Uszkoreit, Jakob and Jones, Llion and Gomez, Aidan N and Kaiser, {\L}ukasz and Polosukhin, Illia},
  journal={Advances in neural information processing systems},
  volume={30},
  year={2017}
}

@article{wang2025endochat,
  title={Endochat: Grounded multimodal large language model for endoscopic surgery},
  author={Wang, Guankun and Bai, Long and Wang, Junyi and Yuan, Kun and Li, Zhen and Jiang, Tianxu and He, Xiting and Wu, Jinlin and Chen, Zhen and Lei, Zhen and others},
  journal={arXiv preprint arXiv:2501.11347},
  year={2025}
}

@article{oord2018representation,
  title={Representation learning with contrastive predictive coding},
  author={Oord, Aaron van den and Li, Yazhe and Vinyals, Oriol},
  journal={arXiv preprint arXiv:1807.03748},
  year={2018}
}

@article{maier2017surgical,
  title={Surgical data science for next-generation interventions},
  author={Maier-Hein, Lena and Vedula, Swaroop S and Speidel, Stefanie and Navab, Nassir and Kikinis, Ron and Park, Adrian and Eisenmann, Matthias and Feussner, Hubertus and Forestier, Germain and Giannarou, Stamatia and others},
  journal={Nature Biomedical Engineering},
  volume={1},
  number={9},
  pages={691--696},
  year={2017},
  publisher={Nature Publishing Group UK London}
}

@inproceedings{li2024mvbench,
  title={Mvbench: A comprehensive multi-modal video understanding benchmark},
  author={Li, Kunchang and Wang, Yali and He, Yinan and Li, Yizhuo and Wang, Yi and Liu, Yi and Wang, Zun and Xu, Jilan and Chen, Guo and Luo, Ping and others},
  booktitle={Proceedings of the IEEE/CVF Conference on Computer Vision and Pattern Recognition},
  pages={22195--22206},
  year={2024}
}

@article{simeoni2025dinov3,
  title={Dinov3},
  author={Sim{\'e}oni, Oriane and Vo, Huy V and Seitzer, Maximilian and Baldassarre, Federico and Oquab, Maxime and Jose, Cijo and Khalidov, Vasil and Szafraniec, Marc and Yi, Seungeun and Ramamonjisoa, Micha{\"e}l and others},
  journal={arXiv preprint arXiv:2508.10104},
  year={2025}
}

@inproceedings{ni2022expanding,
  title={Expanding language-image pretrained models for general video recognition},
  author={Ni, Bolin and Peng, Houwen and Chen, Minghao and Zhang, Songyang and Meng, Gaofeng and Fu, Jianlong and Xiang, Shiming and Ling, Haibin},
  booktitle={European conference on computer vision},
  pages={1--18},
  year={2022},
  organization={Springer}
}

@article{team2024gemma,
  title={Gemma 2: Improving open language models at a practical size},
  author={Team, Gemma and Riviere, Morgane and Pathak, Shreya and Sessa, Pier Giuseppe and Hardin, Cassidy and Bhupatiraju, Surya and Hussenot, L{\'e}onard and Mesnard, Thomas and Shahriari, Bobak and Ram{\'e}, Alexandre and others},
  journal={arXiv preprint arXiv:2408.00118},
  year={2024}
}

@article{dubey2024llama,
  title={The llama 3 herd of models},
  author={Dubey, Abhimanyu and Jauhri, Abhinav and Pandey, Abhinav and Kadian, Abhishek and Al-Dahle, Ahmad and Letman, Aiesha and Mathur, Akhil and Schelten, Alan and Yang, Amy and Fan, Angela and others},
  journal={arXiv e-prints},
  pages={arXiv--2407},
  year={2024}
}

@misc{qwen2025qwen25technicalreport,
      title={Qwen2.5 Technical Report}, 
      author={An Yang and Baosong Yang and Beichen Zhang and Binyuan Hui and Bo Zheng and Bowen Yu and Chengyuan Li and Dayiheng Liu and Fei Huang and Haoran Wei and Huan Lin and Jian Yang and Jianhong Tu and Jianwei Zhang and Jianxin Yang and Jiaxi Yang and Jingren Zhou and Junyang Lin and Kai Dang and Keming Lu and Keqin Bao and Kexin Yang and Le Yu and Mei Li and Mingfeng Xue and Pei Zhang and Qin Zhu and Rui Men and Runji Lin and Tianhao Li and Tianyi Tang and Tingyu Xia and Xingzhang Ren and Xuancheng Ren and Yang Fan and Yang Su and Yichang Zhang and Yu Wan and Yuqiong Liu and Zeyu Cui and Zhenru Zhang and Zihan Qiu},
      year={2025},
      eprint={2412.15115},
      archivePrefix={arXiv},
      primaryClass={cs.CL},
      url={https://arxiv.org/abs/2412.15115}, 
}

@article{gu2021domain,
  title={Domain-specific language model pretraining for biomedical natural language processing},
  author={Gu, Yu and Tinn, Robert and Cheng, Hao and Lucas, Michael and Usuyama, Naoto and Liu, Xiaodong and Naumann, Tristan and Gao, Jianfeng and Poon, Hoifung},
  journal={ACM Transactions on Computing for Healthcare (HEALTH)},
  volume={3},
  number={1},
  pages={1--23},
  year={2021},
  publisher={ACM New York, NY}
}

@inproceedings{song2024moviechat,
  title={Moviechat: From dense token to sparse memory for long video understanding},
  author={Song, Enxin and Chai, Wenhao and Wang, Guanhong and Zhang, Yucheng and Zhou, Haoyang and Wu, Feiyang and Chi, Haozhe and Guo, Xun and Ye, Tian and Zhang, Yanting and others},
  booktitle={Proceedings of the IEEE/CVF Conference on Computer Vision and Pattern Recognition},
  pages={18221--18232},
  year={2024}
}

@article{zhao2025video,
  title={Video-QTR: Query-Driven Temporal Reasoning Framework for Lightweight Video Understanding},
  author={Zhao, Xinkui and Wang, Zuxin and Zhang, Yifan and Cheng, Guanjie and Xu, Yueshen and Deng, Shuiguang and Liu, Chang and Wang, Naibo and Yin, Jianwei},
  journal={arXiv preprint arXiv:2512.09354},
  year={2025}
}

@article{twinanda2016endonet,
  title={Endonet: a deep architecture for recognition tasks on laparoscopic videos},
  author={Twinanda, Andru P and Shehata, Sherif and Mutter, Didier and Marescaux, Jacques and De Mathelin, Michel and Padoy, Nicolas},
  journal={IEEE transactions on medical imaging},
  volume={36},
  number={1},
  pages={86--97},
  year={2016},
  publisher={IEEE}
}

@article{niitsu2013using,
  title={Using the Objective Structured Assessment of Technical Skills (OSATS) global rating scale to evaluate the skills of surgical trainees in the operating room},
  author={Niitsu, Hiroaki and Hirabayashi, Naoki and Yoshimitsu, Masanori and Mimura, Takeshi and Taomoto, Junya and Sugiyama, Yoich and Murakami, Shigeru and Saeki, Shuji and Mukaida, Hidenori and Takiyama, Wataru},
  journal={Surgery today},
  volume={43},
  number={3},
  pages={271--275},
  year={2013},
  publisher={Springer}
}
\end{document}